\documentclass[12pt]{article}
\usepackage[utf8]{inputenc}
\usepackage{microtype}
\usepackage{graphicx}
\usepackage{booktabs} %
\usepackage{natbib}
\usepackage{fancyhdr}

\usepackage[dvipsnames]{xcolor}

\usepackage{hyperref}
\definecolor{mydarkblue}{rgb}{0,0.08,0.45}
\hypersetup{ %
pdftitle={},
pdfauthor={},
pdfsubject={},
pdfkeywords={},
pdfborder=0 0 0,
pdfpagemode=UseNone,
colorlinks=true,
linkcolor=mydarkblue,
citecolor=mydarkblue,
filecolor=mydarkblue,
urlcolor=mydarkblue,
pdfview=FitH}

\usepackage{url}            %
\usepackage{booktabs}       %
\usepackage{amsfonts}       %
\usepackage{nicefrac}       %
\usepackage{microtype}      %

\usepackage{graphicx}
\usepackage{wrapfig}
\usepackage{subcaption}

\usepackage{tablefootnote}
\usepackage{amsmath}
\usepackage{bbold} %
\usepackage{bbm}
\usepackage{amsthm}
\usepackage{txfonts}
\usepackage{mathtools}
\usepackage{graphbox}
\usepackage{xspace}
\usepackage{cleveref}

\DeclareMathOperator*{\argmax}{arg\,max}

\DeclareMathOperator{\opE}{\mathbbm{E}}
\DeclareMathOperator{\opH}{H}
\DeclareMathOperator{\opI}{I}
\DeclareMathOperator{\opp}{p}

\newcommand{\Entropy}[1]{\opH [ #1 ]}

\newcommand{\Hc}[2]{\opH [ #1 \mathbin{\vert} #2 ]}
\newcommand{\MI}[2]{\opI [ #1 ; #2 ]}
\newcommand{\MIt}[3]{\opI [ #1 ; #2 ; #3 ]}
\newcommand{\MIc}[3]{\opI [ #1 ; #2 \mathbin{\vert} #3 ]}
\newcommand{\MIct}[4]{\opI [ #1 ; #2 ; #3 \mathbin{\vert} #4 ]}
\newcommand{\prob}[1]{\opp ( #1 )}
\newcommand{\probc}[2]{\opp ( #1 \mathbin{\vert} #2 )}

\newcommand{\chainedE}[2]{\opE_{#2} {#1}}

\newcommand{\poolx}{(x)_\text{pool}}
\newcommand{\pooly}{(y)_\text{pool}}
\newcommand{\pooldataset}{D_\text{pool}}

\newcommand{\batchx}{(x)_\text{batch}}
\newcommand{\batchy}{(y)_\text{batch}}

\newcommand{\traindataset}{D_\text{train}}

\newcommand{\targetx}{(x)_{\text{eval}}}
\newcommand{\targety}{(y)_{\text{eval}}}
\newcommand{\targetdataset}{D_{\text{eval}}}

\newcommand{\testdataset}{D_{\text{test}}}

\newcommand{\w}{\omega}

\allowdisplaybreaks

\title{PowerEvaluationBALD: Efficient Evaluation-Oriented Deep (Bayesian) Active Learning with Stochastic Acquisition Functions}
\author{%
    Andreas Kirsch\footnote{Correspondence to: \texttt{andreas.kirsch@cs.ox.ac.uk}} \hspace{5mm} %
    Yarin Gal %
    \\
    OATML \\
    Department of Computer Science \\
    University of Oxford\\
    \\
    Report
}

\begin{document}

\maketitle

\begin{abstract}
    We develop BatchEvaluationBALD, a new acquisition function for deep Bayesian active learning, as an expansion of BatchBALD that takes into account an evaluation set of unlabeled data, for example the pool set. We also develop a variant for the non-Bayesian setting, which we call Evaluation Information Gain. To reduce computational requirements and allow these methods to scale to larger acquisition batch sizes, we introduce stochastic acquisition functions that use importance sampling of tempered acquisition scores. We call this method PowerEvaluationBALD. We show in a few initial experiments that PowerEvaluationBALD works on par with BatchEvaluationBALD, which outperforms BatchBALD on Repeated MNIST (MNISTx2), while massively reducing the computational requirements compared to BatchBALD or BatchEvaluationBALD.
\end{abstract}

\section{Introduction}

Active learning is essential for increasing label- and thus cost-efficiency in real-world machine learning applications, especially when they use deep learning, while quantifying uncertainty is important for safety-critical systems. Combining active learning with Bayesian methods for deep neural networks has been an important research avenue for this reason.

In \emph{active learning}, we have access to a huge reservoir of unlabelled data in a pool set. An \emph{acquisition function} selects samples from this pool set to be labeled by an oracle (e.g., a human expert). Ideally, the selected samples increase the performance of the machine learning model faster than random labeling of samples.
Active learning is a necessary component when bootstrapping machine learning solutions, which can be combined with semi-supervised or unsupervised methods to increase label efficiency, or with other methods like the introduction of additional inductive biases to increase data-efficiency\footnote{Data-efficiency refers to enabling higher model performance with less data overall, whereas label-efficiency refers to enabling higher model performance with a similar amount of unlabeled data but fewer labels.}
overall.

\emph{Bayesian neural networks} treat the model parameters $\w$ as a distribution $\prob{\w}$. Using training data $\traindataset$, a posterior distribution $\probc{\w}{\traindataset}$ is inferred. The posterior model uncertainty $\Hc{\w}{\traindataset}$ is reduced using more training data $\traindataset$. This contrasts with regular deep learning, which only learns a maximum-likelihood point estimate of the model parameters.
The posterior parameter distribution $\probc{\w}{\traindataset}$ induces a distribution over predictions for a given sample $\probc{y}{x,\w}$, which allows for uncertainty quantification: via marginalization $\probc{y}{x,\traindataset}$ to obtain the aleatoric uncertainty $\Hc{Y}{x,\traindataset}$, and via the expectation over $\probc{\w}{\traindataset}$ to obtain the epistemic uncertainty $\Hc{Y}{x,\traindataset} - \chainedE{\Hc{Y}{x,\w}}{\probc{\w}{\traindataset}}=\MIc{Y}{\omega}{\traindataset}$.

Even though much progress has been made in approximating Bayesian inference efficiently and at scale, uptake has been slow, and while Bayesian active learning has shown promise, it is still not as widely adapted as its principled foundations would suggest.

We hope this report can contribute to increasing this uptake by contributing: 
\begin{enumerate}
    \item a new Bayesian acquisition function based on information-theoretic principles which identifies the most informative samples in regards to the distribution of an evaluation set of the unlabeled data;
    \item stochastic acquisition functions which allow for acquiring diverse batches of samples with less computational effort than alternative methods; and
    \item a new experimental protocol that is more aligned with real-word applications by including data duplication, noisy oracles, and class imbalances into the evaluation.
\end{enumerate}
As this is a preprint, we provide simple experiments for the main claims, while outlining future experiments, ablations, and directions.

Usually, batches of samples are acquired to avoid retraining models after each individual acquisition. However, most existing acquisition functions score points individually, and batches are constructed using these individual scores.
This leads to sub-optimal behavior. To acquire a batch, the highest scoring samples are selected, which may result in a lack of diversity: the samples that are the most informative for a given model at a certain time might be quite similar, and thus not as informative jointly as a more diverse batch. Batches need to be viewed holistically because the acquisition functions that score samples individually can be myopic. Batch acquisition functions fix this.

Within Bayesian Deep Learning, BatchBALD \citep{kirsch2019batchbald} was one of the first to introduce the concept of batch acquisition functions that score potential candidate batches jointly. Using information-theoretic principles, BatchBALD expands on BALD \citep{houlsby2011bayesian}, which estimates the information gain between a candidate sample and the Bayesian model parameters, the mutual information $\MI{y}{\w}$, to take into account the information gain of a batch of candidate samples with the parameters, the joint mutual information $\MI{\batchy}{\w}$. Importantly, the joint mutual information in BatchBALD is sub-modular, which allows for a greedy algorithm to grow batches while guaranteeing $1-\tfrac{1}{e}$-optimality.
Even though BatchBALD performs very well for small acquisition batch sizes and picks batches with higher diversity than BALD, it is computationally challenging as the entropies of larger joint distributions need to be evaluated. BatchBALD suffers severely from the combinatorial explosion for acquisition batch sizes of just 10 samples already.
While approximations help, the quality of the joint entropy estimates quickly degrades, leading to later samples in a batch being picked at random essentially.
This poses an issue for applications that require larger acquisition batch sizes.

Another promising approach consists of Coresets, which we need to explain in more detail but will not (yet).

\section{Background}

\subsection{Problem Setting}

In a Bayesian setting, the model parameters of our Bayesian model $M$ are a random variable $\w \sim \probc{\w}{\traindataset}$, which we sample from the posterior distribution. To sample predictions $y$, we integrate out the model parameters $y \sim \probc{y}{x, \traindataset} = \chainedE{\probc{y}{x, \w}}{\probc{\w}{\traindataset}}$.

In active learning, we start with an unlabeled \emph{pool set} $\pooldataset$ and a \emph{training set} $\traindataset$, which can either be empty or consist of an initial set of labeled samples. Given the training set, we train a model that can make predictions $\probc{y}{x, \traindataset}$, where conditioning on the training set signifies that we have trained a model with parameters $\w$ on it, and $x$ and $y$ are the input sample and its prediction respectively.
We also have access to an \emph{oracle} that can tell us the correct label $\hat{y}$ for a given sample $x \in \pooldataset$, which we can then move from the pool set to the training set before training a new model.
The goal in active learning is to achieve a certain prediction accuracy on the \emph{test set} $\testdataset$ with the smallest number of queried labels.

As we do not want to train a new model after each individual query, we query the labels for a batch of samples before retraining. To select a batch, we use a batch acquisition function \citep{kirsch2019batchbald} $a \left (\batchx, \probc{\w}{\traindataset} \right )$ that jointly scores a candidate batch  $\batchx = \left \{ x_1, \ldots, x_B \right \}$. We want to find the optimal candidate batch $\batchx^* = \left \{ x^*_1, \ldots, x^*_B \right \}$ for a fixed acquisition batch size $B$:
\begin{align}
    \batchx^* = \argmax_{\batchx \in \pooldataset} a \left (\batchx, \probc{\w}{\traindataset} \right ).
\end{align}

\subsection{Additional Notation}

While $x$ and $y$ denote the otherwise unspecified samples and predictions, we denote the data and predictions of the unlabeled pool $\poolx$ and $\pooly$. Similarly, $\batchx$ and $\batchy$ are the data and predictions of the current batch candidates. Moreover, we let $\targetdataset$, made up of data $\targetx$ and predictions $\targety$ denote a dataset which is sampled from the data distribution that is used for evaluation, for example the test or validation set---more about this in \cref{sec:eal}.

\subsection{BALD and BatchBALD}

BatchBALD \citep{kirsch2019batchbald} is the mutual information between the joint of the batch candidate's predictions and the model parameters:
\begin{align}
    a_\text{BatchBALD} \left (\batchx, \probc{\w}{\traindataset} \right ) := \MIc{\batchy}{\w}{\batchx, \traindataset}. 
\end{align}
We can view BALD \citep{houlsby2011bayesian} as an upper-bound of BatchBALD
\begin{align}
    a_\text{BatchBALD} \left (\batchx, \probc{\w}{\traindataset} \right ) &\le a_\text{BALD} \left (\batchx, \probc{\w}{\traindataset} \right ) \\
    &:= \sum_{b=1}^B \MIc{y_b}{\w}{x_b, \traindataset},
\end{align}
with equality for individual acquisitions (acquisition batch size 1).
$a_\text{BALD}$ can be maximized by computing individual scores $\MIc{y}{\w}{x, \traindataset}$ on the pool set and selecting the $B$ highest scoring samples.

\section{Real-world datasets}
\label{sec:rwd}

Compared to datasets that are commonly used in machine learning applications like MNIST or CIFAR-10, real-world unlabeled datasets are not well curated. For example, they suffer from:
\begin{enumerate}
    \item redundant or duplicated data; \label{enum:rwd:data_duplication}
    \item noisy oracles that lead to corrupted or noisy labels; \label{enum:rwd:noisy_labels}
    \item class imbalances; \label{enum:rwd:class_imbalances}
    \item heteroscedastic data noise; and \label{enum:rwd:data_noise}
    \item outliers or out-of-distribution data that was not filtered out properly. \label{enum:rwd:ood_data}
\end{enumerate}
To benchmark active learning methods, we want to move away from well-curated datasets and instead focus more on datasets that are flawed and resemble what we would find in the real world.
\citet{kirsch2019batchbald} introduced RMNIST (Repeated MNIST), which covers \cref{enum:rwd:data_duplication}: it contains a parameterized number of MNIST copies with added Gaussian noise to make all samples unique. It clearly showed that BALD was not coping well with redundant data and would generally pick batches that were self-similar.

This points towards another issue: in an active learning setting, the unlabeled set might not match the test and validation set's distribution due to issues 1.--4. The unlabeled dataset might itself be partially out-of-distribution compared to the test set distribution. Acquisition functions do not take this into account. Indeed, acquisition functions like BALD are agnostic of the test set and pool set distributions, and the sampled training set will be biased.

\subsection{Remark on AI Fairness and Class Imbalances}

An example for a distribution mismatch between the unlabeled dataset and the test dataset is heavy class imbalances in the collected data. This is also an issue that is relevant to AI fairness, where one wants to ensure that the selected samples follow a balanced distribution that might be different from the collected pool set distribution.

\section{Evaluation-Oriented Active Learning}
\label{sec:eal}

To take the distribution of the test set or an validation set into account, an acquisition function should ideally select samples that are informative towards this set, which represents the data distribution we are truly interested in. 
This motivation is similar to \citet{jain2020information}, which state as part of their motivation for an otherwise unrelated implementation:
\begin{quote}
    This motivates our choice of acquisition function as one that selects the set of points whose acquisition would maximize the information gained about predictive distribution on the unlabeled set.
\end{quote}

\subsection{Evaluation Information Gain}

\newcommand{\testinfogain}{EIG\xspace}

While (Batch-)BALD measures the information gain for the model parameters given the candidate batch, which we can write informally as:
\begin{align}
    \Entropy{\w} - \Hc{\w}{\batchy} = \MI{\w}{\batchy} = a_\text{BatchBALD} \left (\batchx, \probc{\w}{\traindataset} \right ),
\end{align}
we want to measure the information gain on the evaluation set given the candidate batch. Informally,
\begin{align}
    a_\text{BatchEIG} \left (\batchx, \probc{\w}{\traindataset} \right ) := \Entropy{\targety} - \Hc{\targety}{\batchy} = \MI{\targety}{\batchy}.
\end{align}
We call this new acquisition function (Batch) Evaluation Information Gain to contrast it with the information gain that BALD represents. Similar to BatchBALD and BALD, we define an approximation EIG with an acquisition function $a_\text{EIG}$ that scores points individually.

Importantly, the Evaluation Information Gain does not depend on the model parameters $\w$ as they are marginalized out. It can thus be computed using maximum-likelihood point estimates of the model parameters, too.
BatchEIG is only meaningful when we can condition on $\w$ as we cannot capture the joints distributions otherwise.

\subsection{Evaluating the Evaluation Information Gain}

To see how we can compute BatchEIG, we write it down formally by including the input samples that correspond to the predictions and the training set, and use the symmetry of the mutual information:
\begin{align*}
    &a_\text{EIG} \left (\batchx, \probc{\w}{\traindataset} \right ) 
    = \MI{\targety}{\batchy} 
    = \MI{\batchy}{\targety} \\
    &= \Hc{\batchy}{\batchx, \traindataset} 
    -
    \Hc{\batchy}{\batchx, \targety, \targetx, \traindataset} \\
    &\approx 
    \Hc{\batchy}{\batchx, \traindataset} - \Hc{\batchy}{\batchx, \targetdataset, \traindataset} \\
    &=
    \Hc{\batchy}{\batchx, \traindataset} - \Hc{\batchy}{\batchx, \targetdataset \cup \traindataset}.
\end{align*}
The LHS term is the predictive (joint) entropy of the batch samples given a model trained on the training set;
the RHS term is the predictive (joint) entropy of the batch samples given a model trained on the training set and evaluation set.
Picking the test set as the evaluation set is commonly frowned upon---instead we could choose a separate validation set for this purpose and only use the test set for independent performance evaluation as is custom.

\subsection{Choice of the Evaluation Set}
Training on a separate validation set requires labelling additional samples that are held back, however. For comparison with other active learning methods, these samples would have to be counted towards the total number of acquired samples. This would severely handicap this proposed method. Even though deep active learning experiments usually use a validation set for hyperparameter tuning, it is an acknowledged issue that widens the gap between research and real-world applications. We do not want to expand this gap.

For this reason, we choose another option for better comparability: assuming as usual that we do not suffer from a distribution shift between the test set and the unlabeled pool set, we choose the pool set as evaluation set $\targetdataset = \pooldataset$. While we do not know the labels, we can just use the predictions from the model trained on $\traindataset$ itself. We leave using semi-supervised methods like pseudo-labels for future analysis and instead use \emph{self-distillation}.

However, this also means that $\batchx \subseteq \targetx = \poolx$, which is problematic as this ought to reduce the epistemic uncertainty to 0 for all batch candidates on the RHS. To avoid this, we could subsample the pool set to build the evaluation set and make sure that we do not pick batch candidates from it. In practice, we have found that using the pool set is sufficient, and the model does not overfit. We also leave this for future explorations.

\subsection{EvaluationBALD}

\begin{figure}[th]
    \centering
    \includegraphics[width=0.6\linewidth,clip,trim=15 12 32 8]{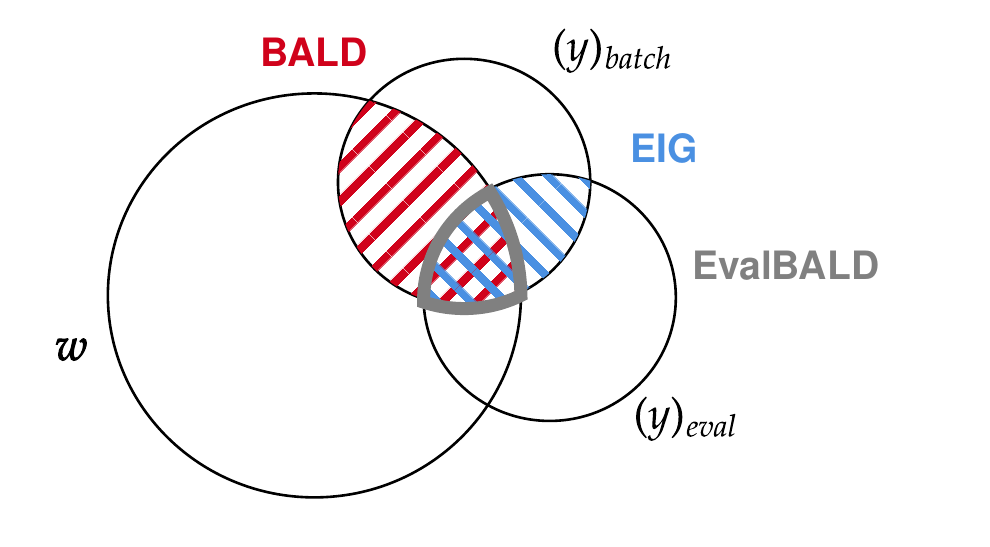}
    \caption{\emph{I-diagram depicting $\w$, $\pooly$ and $\batchy$}. $\poolx$, $\batchx$ and $\traindataset$ are not depicted. The \emph{(Batch)BALD} term is depicted in red. The \emph{Evaluation Information Gain} term is depicted in blue. The ``combined'' \emph{EvaluationBALD} term is depicted in gray.
    }
    \label{fig:idiagram}
\end{figure}

We can expand \testinfogain to the Bayesian setting by comparing it to BatchBALD. We can depict both approaches in an I-diagram, see \cref{fig:idiagram}.
We are interested in the mutual information between the predictions for the batch candidates and the evaluation set \emph{and} the model parameters. Formally:
\begin{align}
    a_\text{EvalBALD} \left (\batchx, \probc{\w}{\traindataset} \right ) := \MIct{\batchy}{\targety}{\w}{\batchx, \targetx, \traindataset}.
\end{align}
A triple mutual information can be negative. This suits the intuition that some samples in the pool set might actually hinder training. (Batch)BALD cannot capture this.

\subsubsection{Evaluating EvaluationBALD}

Just as we expanded the dual mutual information term into an entropy and conditional entropy before, we can expand the triple mutual information into a dual mutual information and a dual conditional mutual information. Using $\MIt{x}{y}{z}=\MI{x}{y} - \MIc{x}{y}{z}$, we obtain:
\begin{align*}
    & a_\text{EvalBALD} \left (\batchx, \probc{\w}{\traindataset} \right ) = \MIct{\batchy}{\targety}{\w}{\batchx, \targetx, \traindataset} \\
    & = \MIc{\batchy}{\w}{\batchx, \targetx, \traindataset} - \MIc{\batchy}{\w}{\batchx, \targety, \targetx, \traindataset} \\
    & = \MIc{\batchy}{\w}{\batchx, \traindataset} - \MIc{\batchy}{\w}{\batchx, \targety, \targetx, \traindataset} \\
    & \approx \MIc{\batchy}{\w}{\batchx, \traindataset} - \MIc{\batchy}{\w}{\batchx, \targetdataset \cup \traindataset} \\
    & = a_\text{BatchBALD} \left (\batchx, \probc{\w}{\traindataset} \right ) -
    a_\text{BatchBALD} \left (\batchx, \probc{\w}{\targetdataset \cup \traindataset} \right ).
\end{align*}
This formulation provides a nice intuition: 
\begin{quote}
    A sample will obtain a high EvaluationBALD score when the LHS term is high and the RHS term is low. This is true for samples that have high epistemic uncertainty given just the training set, but have low epistemic uncertainty given the combined dataset. They cannot be easily explained given the current data, but they lie in dense areas of the evaluation set (combined with the training set).
    
    On the other hand, a sample will obtain a low EvaluationBALD score, when the uncertainty increases in the RHS compared to LHS. This is most likely for samples that contradict the evaluation set.\footnote{Given that we use self-distillation, it seems unlikely uncertainty on the RHS can increase over the LHS.}
\end{quote}
This leaves us with a simple way to compute EvaluationBALD via the difference of two BatchBALD terms. Albeit, it is quite computationally intensive and does not scale easily to large acquisition batch sizes. This might prevent using EvaluationBALD in certain applications.

\section{Stochastic Acquisition Functions}

BatchBALD performs better than BALD because it scores batch candidates jointly and takes into account the redundancy between them. It correctly scores a more diverse batch higher than a batch of similar points. A computationally cheaper way of achieving diversity is by adding noise. Indeed, BALD performs better with MC Dropout with fewer MC samples, whereas BatchBALD requires a high number of MC Dropout samples.

We propose stochastic acquisition functions as a way to increase diversity while scoring points individually.
For this, we treat the scores $a_p := a \left( x_p, \probc{\w}{\traindataset} \right )$ for $x_p \in \pooldataset$ as an unnormalized probability distribution. To focus on higher scorers, we take it to the power $\alpha$ and sample the batch without replacement using the following probability distribution:
\begin{align}
    p(x_p) = \frac{a_p^\alpha}{\sum_{x_q \in \pooldataset} a_q^\alpha}.
\end{align}
We have found this to perform very well empirically for small $\alpha (\approx 5)$. 

We can combine this approach with nonbatch acquisition functions. This leads to stochastic acquisition methods that we call PowerBALD, PowerEvaluationBALD, and PowerEIG.

Of additional note is that for $\alpha=0$, we obtain random acquisition. This allows for smooth interpolation between random acquisition and other acquisition functions. Annealing $\alpha$ could especially help with cold-started active learning (in which there is no initial training set). 

On the other hand, large $\alpha$ will lead to a loss of stochasticity, and the batch samples will almost surely be picked in the order of their acquisition scores.

\section{Experiments}

\begin{figure}[th]
    \centering
    \includegraphics[width=1\linewidth]{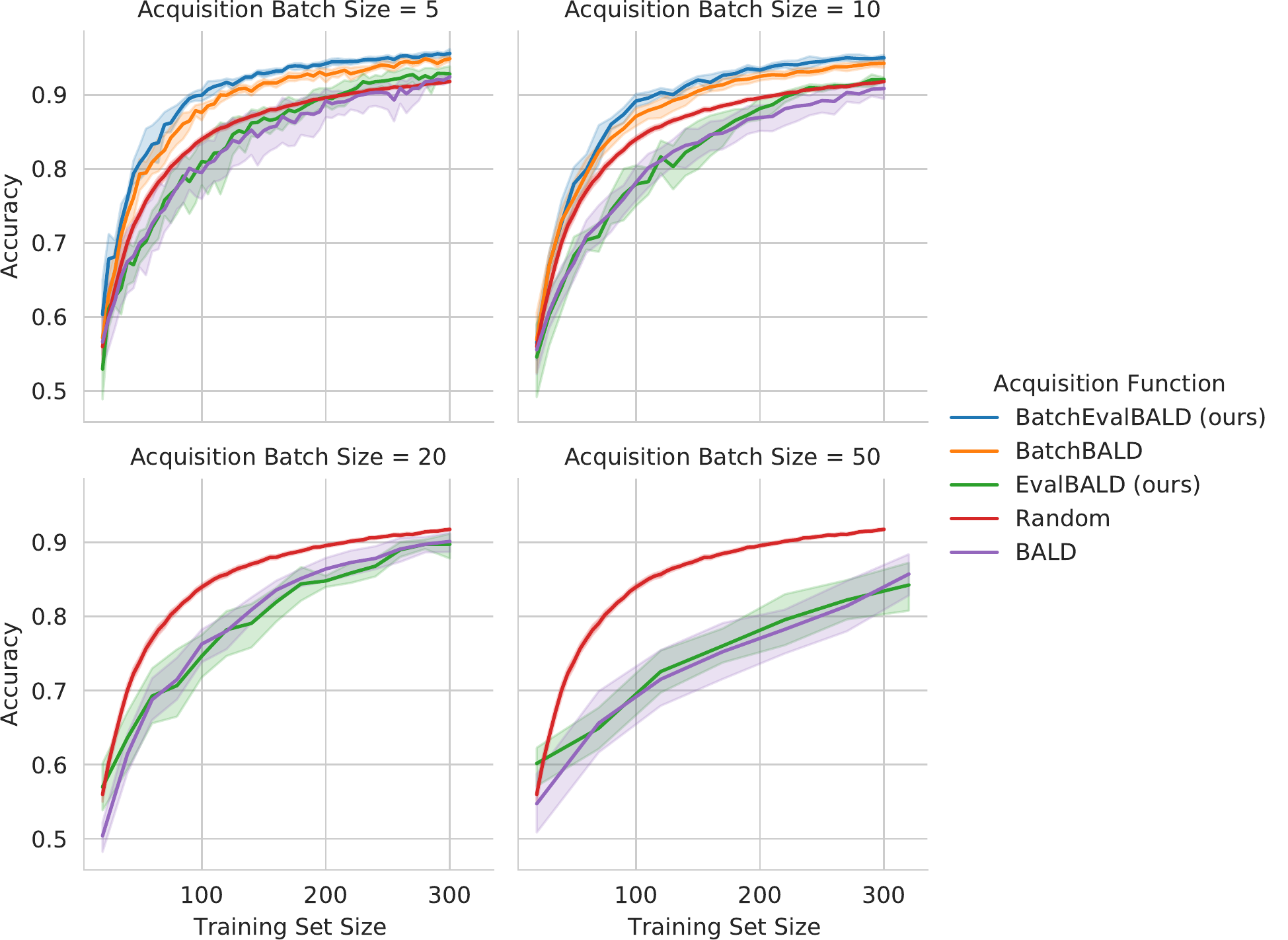}
    \caption{\emph{Performance on \emph{RMNIST=MNISTx2} for increasing training set size using batch acquisitions (BatchEvaluationBALD, BatchBALD) or their approximations (EvaluationBALD, BALD).} BatchEvaluationBALD outperforms BatchBALD.
    }
    \label{fig:evalbald_vs_bald}
\end{figure}

\begin{figure}[th]
    \centering
    \includegraphics[width=1\linewidth]{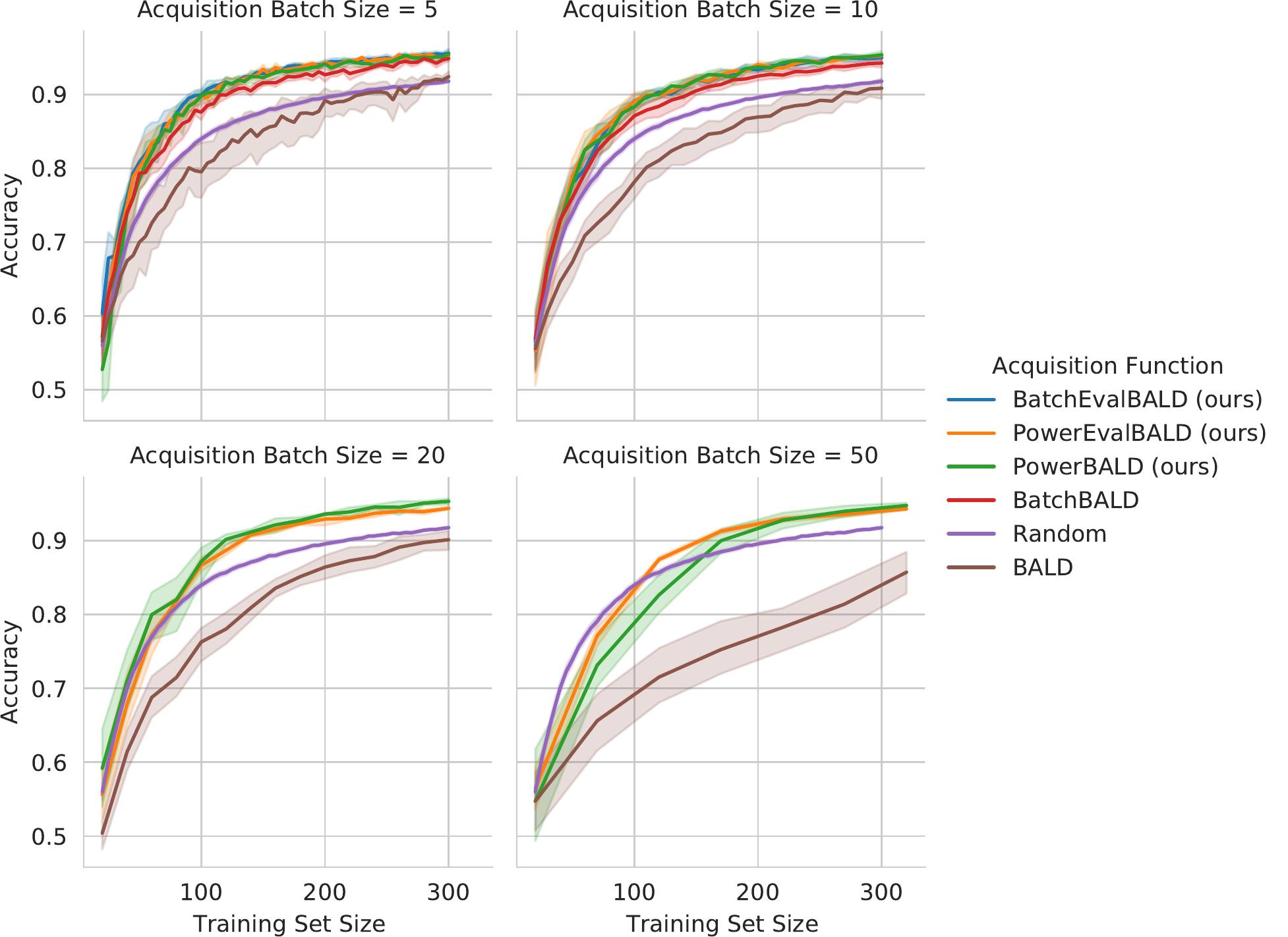}
    \caption{\emph{Performance on \emph{RMNIST=MNISTx2} for increasing training set size using stochastic acquisition functions (PowerEvaluationBALD, PowerBALD) or the deterministic batch  variants (BatchEvaluationBALD, BatchBALD).} The stochastic acquisition functions perform on par with the computationally more expensive batch variants. PowerEvaluationBALD and PowerBALD perform on par with BatchEvaluationBALD and are better than BatchBALD.
    }
    \label{fig:power_vs_batch}
\end{figure}

\begin{figure}[th]
    \centering
    \includegraphics[width=1\linewidth]{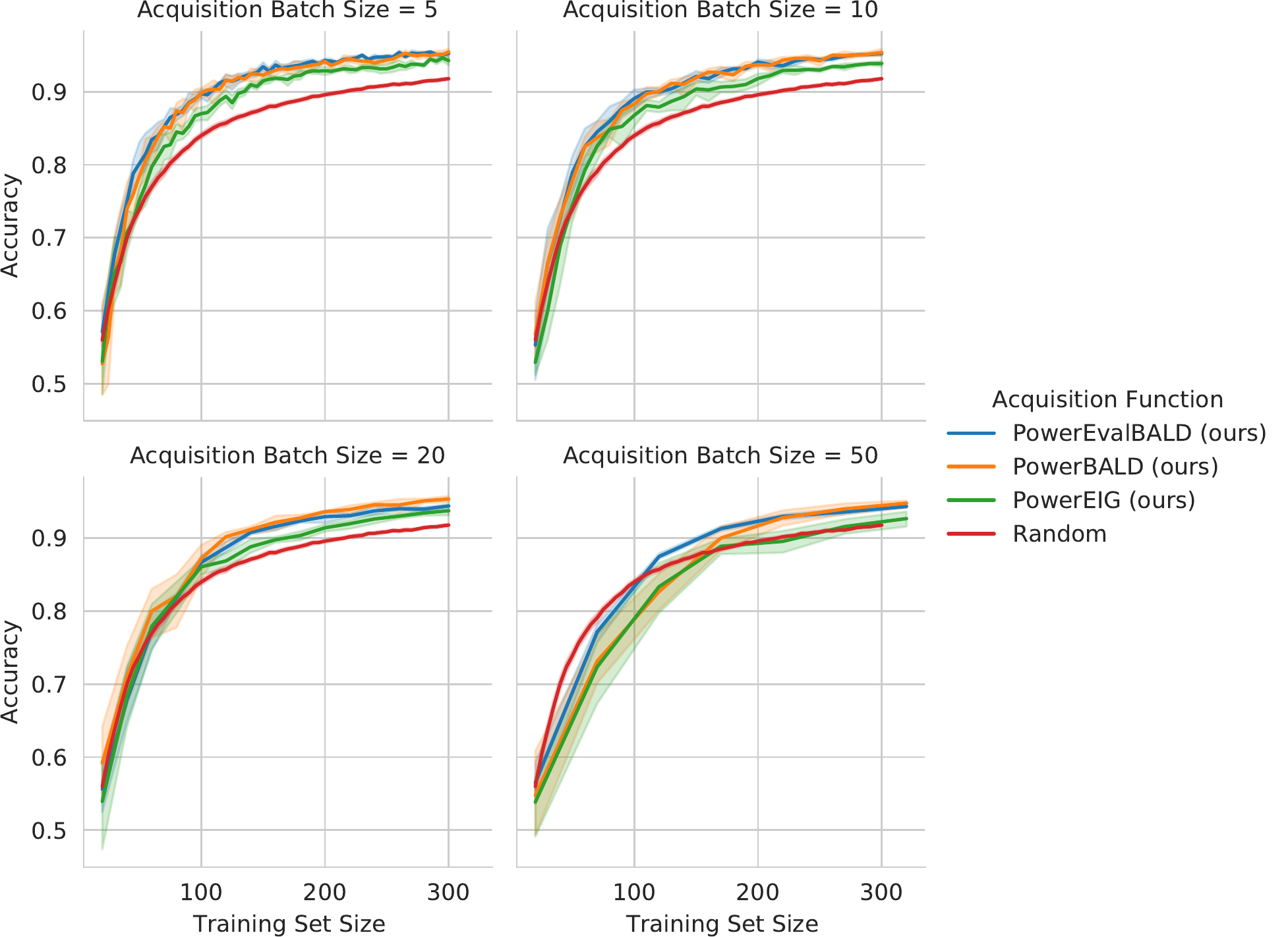}
    \caption{\emph{Performance on \emph{RMNIST=MNISTx2} for increasing training set size using stochastic acquisition functions (PowerEvaluationBALD, PowerBALD, and PowerEIG).} PowerEvaluationBALD performs better than PowerBALD for higher acquisition batch sizes, which in turn performs better than PowerEIG.
    }
    \label{fig:power_eval_vs_bald_vs_eig}
\end{figure}

\begin{figure}[th]
    \centering
    \includegraphics[width=1\linewidth]{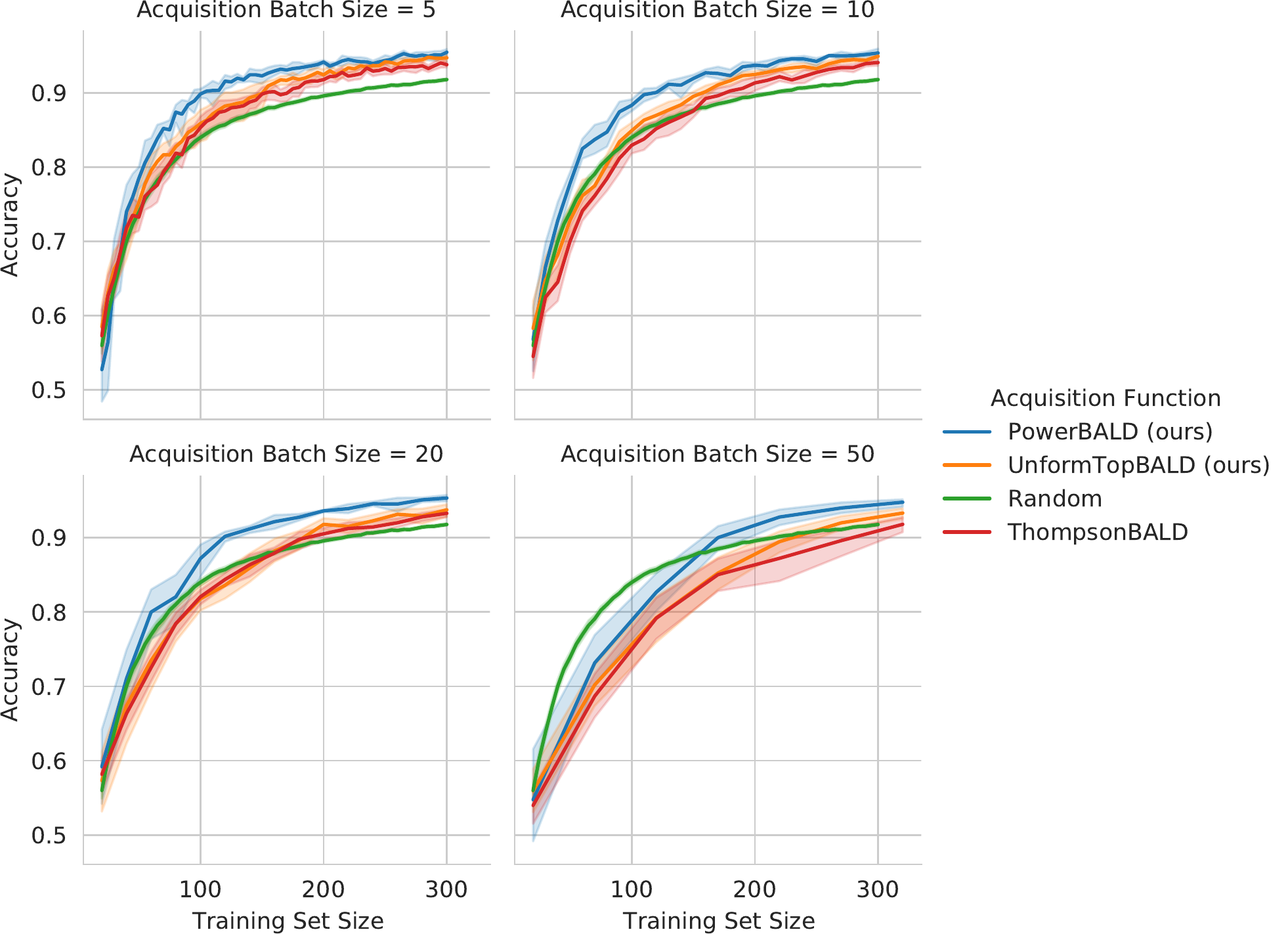}
    \caption{\emph{Performance on \emph{RMNIST=MNISTx2} for increasing training set size using stochastic acquisition functions (PowerBALD, ThompsonBALD, and UniformTopBALD).} UniformTopBALD select $B$ samples uniformly out of the $B*C$ top scorers of an acquisition function (in this case BALD). PowerBALD performs better than UniformTopBALD, which in turn performs better than ThompsonBALD for higher acquisition batch sizes.
    }
    \label{fig:stochasticity_comparison}
\end{figure}

\subsection{Setup}
We use an experimental setting similar to \citet{kirsch2019batchbald}. Bayesian neural networks are approximated via MC dropout. Due to time constraints, we only provide experiments on $RMNIST=MNIST\times2$ which covers \cref{enum:rwd:data_duplication} and \cref{enum:rwd:data_noise} from \cref{sec:rwd}. Due to an oversight, we have oversampled the training set to 24k samples instead of 5096 (which is used in BatchBALD\footnote{Which in turn was picked using grad-student descent.}), which led to worse performance. The results in this report cannot be compared to \citet{kirsch2019batchbald} directly. Rerunning and expanding the experiments is left as future work. All experiments are run at least 5 times and 95\% confidence intervals are reported. For Batch variants, 100 MC dropout samples were used. For other variants, 20 MC dropout samples were used. We set $\alpha = 8$ for all stochastic acquisition functions ($\alpha = 5$).

\subsubsection{Self-distillation}

We perform self-distillation by training a new model on the predictions of the parent model with a KL divergence loss.

\subsection{Experiment Results}

Overall, BatchEvaluationBALD performs better than BatchBALD, see \cref{fig:evalbald_vs_bald}. PowerEvaluationBALD and PowerBALD perform on par with BatchEvaluationBALD, while being much cheaper to evaluate, see \cref{fig:power_vs_batch}. Importantly, PowerBALD performs better than BatchBALD.
PowerEvaluationBALD and PowerBALD perform similarly for small acquisition batch sizes. For larger acquisition batch sizes, PowerEvaluationBALD performs better than PowerBALD. PowerEIG as a non-Bayesian performs the worst of all three. It still performs well compared to BALD and even BatchBALD. Compared to \citet{jeon2020thompson} and another simple stochastic baseline, which samples a batch of size $B$ uniformly from the top $B \times C$ samples (where $C$ is the number of classes), PowerBALD, and thus PowerEvaluationBALD perform better, see \cref{fig:power_eval_vs_bald_vs_eig}.

\section{Conclusion}

We have introduced a new principled acquisition function (Batch)EvaluationBALD that outperforms BatchBALD on Repeated MNIST. Moreover, we have introduced stochastic acquisition functions that allow to smoothly interpolate between random acquisition and deterministic acquisition functions. First experiments show that this added stochasticity helps outperform deterministic methods and is close in performance to computationally costly methods like BatchBALD and BatchEvaluationBALD.
Lastly, we have discussed suggestions for modifying datasets to include challenges that bring them closer to real-world settings, which outlines future experiments.

\newpage
\subsubsection*{Acknowledgments}
We would like to thank Joost van Amersfoort for helpful discussions, feedback, and support; Sebastian Farquhar and Tom Rainforth for helpful discussion and feedback; and all of OATML for providing an amazing environment to conduct research in. AK is supported by the UK EPSRC CDT in Autonomous Intelligent Machines
and Systems (grant reference EP/L015897/1).

\bibliographystyle{plainnat}
\bibliography{power_data_bald}

\end{document}